%% file: ICRA2021_Ossadnik.tex
\documentclass[letterpaper, 10 pt, conference]{ieeeconf}  

\IEEEoverridecommandlockouts                              

\usepackage{amssymb}  
\usepackage{mathtools}

\usepackage{import}
\usepackage{pgfplots}
\usepackage{color}
\usepackage{tabularx} 
\usepackage{multirow}
\usepackage{pgf}

\usepackage{balance}

\newcommand{\mat}[1]{\boldsymbol{#1}}
\newcommand{\vect}[1]{\boldsymbol{#1}}

\title{\LARGE \bf
	Nonlinear stiffness allows passive dynamic hopping for one-legged robots with an upright trunk 
}

\author{Dennis Ossadnik, Elisabeth Jensen and Sami Haddadin$^{*}$
	\thanks{$^{*}$All authors are with the Chair of Robotics Science and Systems Intelligence (RSI), 
		Munich School of Robotics and Machine Intelligence (MSRM),
		Technical University of Munich, 80797 Munich, Germany
		{\tt\small \{dennis.ossadnik, elisabeth.jensen, haddadin\}@tum.de}}%
}

\begin{document}

	\maketitle
	\thispagestyle{empty}
	\pagestyle{empty}

	\begin{abstract}
Template models are frequently used to simplify the control dynamics for robot hopping or running. Passive limit cycles can emerge for such systems and be exploited for energy-efficient control. A grand challenge in locomotion is trunk stabilization when the hip is offset from the center of mass (CoM). The swing phase plays a major role in this process due to the moment of inertia of the leg; however, many template models ignore the leg mass. In this work, the authors consider a robot hopper model (RHM) with a rigid trunk and leg plus a hip that is displaced from the CoM. It has been previously shown that no passive limit cycle exists for such a model given a linear hip spring. In this work, we show that passive limit cycles can be found when a nonlinear hip spring is used instead. To the authors' knowledge, this is the first time that a passive limit cycle has been found for this type of system.
	\end{abstract}

\input{intro}

\input{methods}

\section{Results}
\begin{figure}
	\import{./python/}{limit_cycles.pgf}
	\vspace{-1cm}
	\caption{Phase portrait for the trunk inclination angle and angular velocity (above) and for the leg angle and angular velocity (below). Three gait cycles were completed before falling.}\label{fig:phase}
\end{figure}
\begin{figure*} [h]
	\vspace{1cm}
	\includegraphics[width=0.33\textwidth]{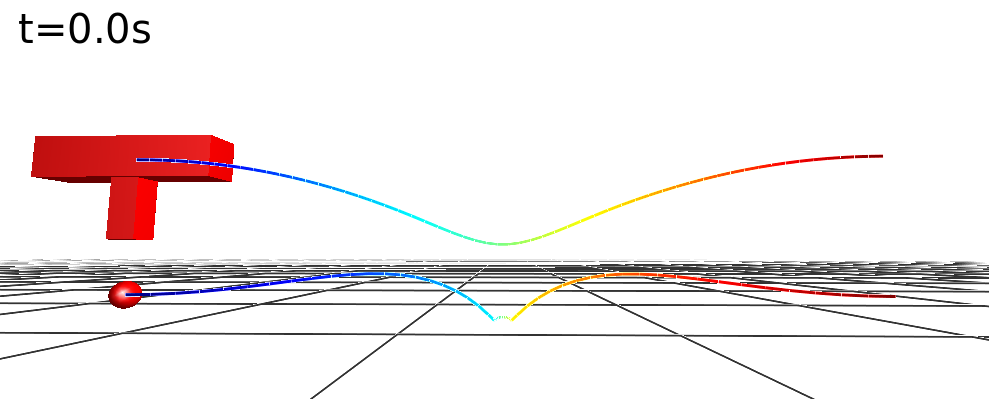}
	\includegraphics[width=0.33\textwidth]{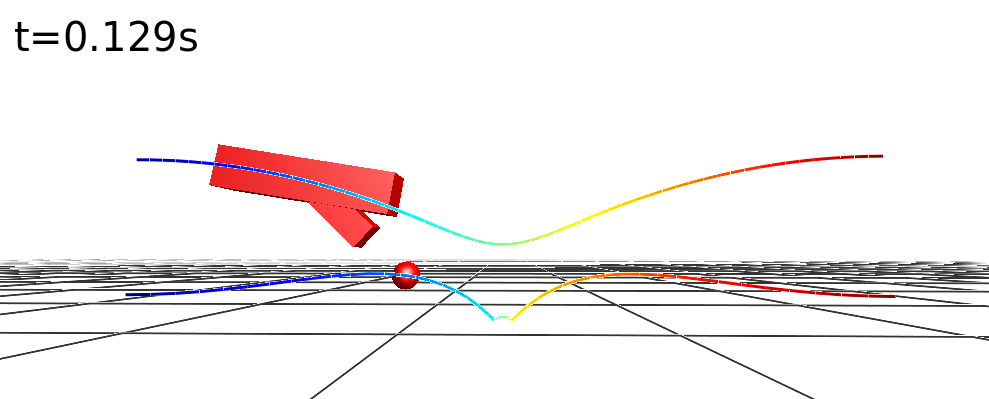}
	\includegraphics[width=0.33\textwidth]{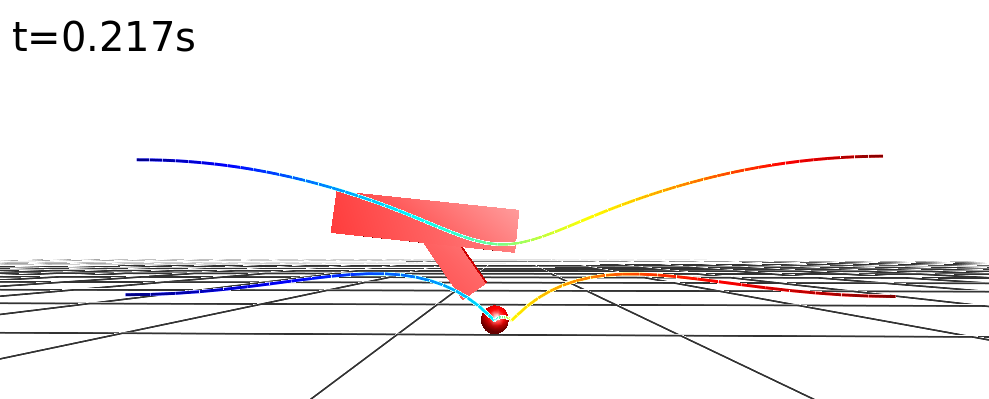}
	\includegraphics[width=0.33\textwidth]{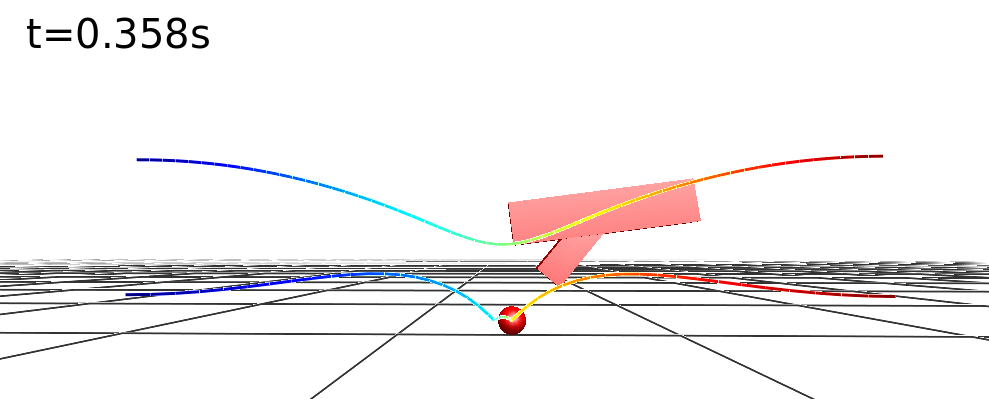}
	\includegraphics[width=0.33\textwidth]{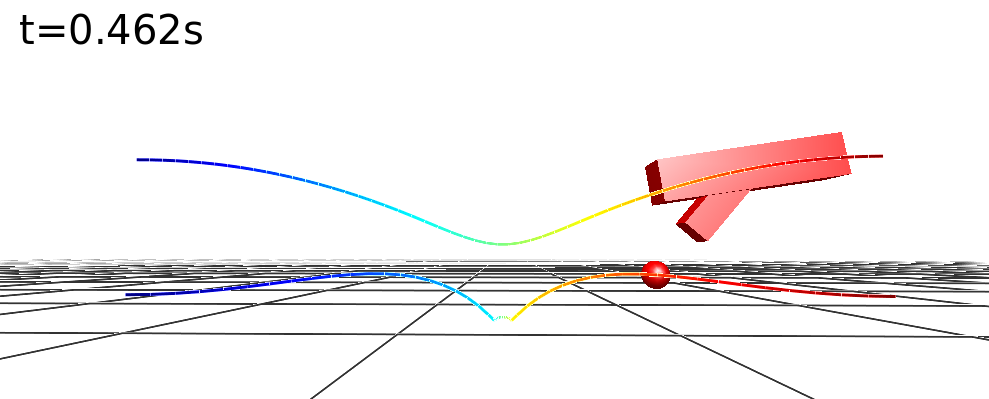}
	\includegraphics[width=0.33\textwidth]{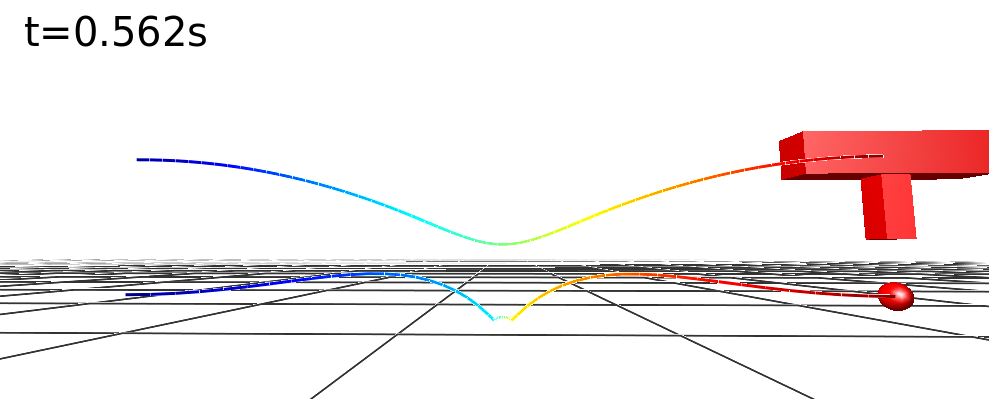}
		\caption{Visualization of limit cycle jumping for the centered-hip cubic-spring (CHCS) model. The images were rendered with the open-source software MeshUp (https://github.com/ORB-HD/MeshUp).} \label{fig:Hyon}
\end{figure*}

A passive limit cycle was found for the centered-hip linear-spring (CHLS) model with identical parameters to those described by \cite{hyon2004passive} (see Table \ref{tbl:values}) and a fixed forward velocity of $\dot{x}= 5 \mathrm{m}/\mathrm{s}$. The fixed point and the eigenvalues of the monodromy matrix can be found in Tables \ref{tbl:fixpt} and \ref{tbl:eig}. The eigenvalues of parameters $z$ and ${\theta}$ are $-5.775$ and $2.243$, respectively, indicating an unstable limit cycle. The simulation was able to complete 5 steps before failure. The change in kinetic energy over the gait cycle was found to be $-0.012$$J$. The associated GRF vectors can be viewed in Fig. \ref{fig:GRF} and a visualization of this gait can be seen in Fig. \ref{fig:Hyon}.

\begin{table}
	\vspace{0.25cm}
	\centering
	\caption{Fixed points for centered-hip linear-spring, centered-hip cubic-spring, upright-trunk cubic-spring, and upright-trunk exponential models. The values correspond to the state vector, $\vect{x} = [x, z, \theta, \phi, l, \dot{x}, \dot{z}, \dot{\theta}, \dot{\phi}, \dot{l}]^{\mathsf{T}}$.} \label{tbl:fixpt}
	\begin{tabularx}{0.48\textwidth}{c|c|c|c}
		CHLS & CHCS & UTCS & UTES \\
		\hline
		& & & \\
		$ \begin{bmatrix} 0.00 \\
		0.637\\
		9.91e-08\\
		-1.12e-07\\
		0.500\\
		5.00\\
		0.000\\
		-1.78\\
		10.5\\
		0.00
		\end{bmatrix} $  & $ \begin{bmatrix}
		0.00\\
		0.579\\
		1.75e-06\\
		-8.05e-07\\
		0.500\\
		5.00\\
		0.00\\
		-1.59\\
		9.45\\
		0.00 \end{bmatrix} $ & $ \begin{bmatrix}
		0.00\\
		1.00\\
		6.46e-06\\
		5.89e-06\\
		0.500\\
		5.00\\
		0.00\\
		-0.587\\
		6.83\\
		0.00 \end{bmatrix} $ & $ \begin{bmatrix}
		0.00\\
		1.24\\
		3.34\\
		3.32\\
		0.500\\
		5.00\\
		0.00\\
		-0.538\\
		6.39\\
		0.00 \end{bmatrix} $
	\end{tabularx}		
\end{table}

\begin{table}
	\centering
	\caption{Eigenvalues for each model. The variables correspond to the reduced state vector, $\vect{x}_{red} = [z, \theta, \phi, \dot{x}, \dot{\theta}, \dot{\phi}]$.} \label{tbl:eig}
	\begin{tabularx}{0.48\textwidth}{c|c|c|c}
		CHLS & CHCS & UTCS & UTES \\
		\hline
		& & & \\
		$ \begin{bmatrix} 
		-5.78 \\
		2.24\\
		1.00\\
		-0.187\\
		-0.0511\\
		-1.23e-05\\
		\end{bmatrix}$ &$ \begin{bmatrix}
		-3.48\\
		1.62\\
		1.00\\
		0.0485\\
		3.02e-06\\
		-0.773 \end{bmatrix}$ & $\begin{bmatrix}
		33.5\\
		-9.42\\
		1.00\\
		0.117\\
		0.0183\\
		-0.000316\\
		\end{bmatrix}$ & $\begin{bmatrix}
		-56.1\\
		22.8\\
		0.999\\
		-0.0298\\
		-0.0117\\
		0.000564\\
		\end{bmatrix}$   
	\end{tabularx}		
\end{table}


\begin{figure}[h]
	\import{./python/}{GRF_neu.pgf}
	\caption{Normalized ground reaction force vectors in the CoM frame of reference. In all cases the vectors intersect well above the CoM.}\label{fig:GRF}
\end{figure}

	\begin{figure*}  [h]
	\includegraphics[width=0.33\textwidth]{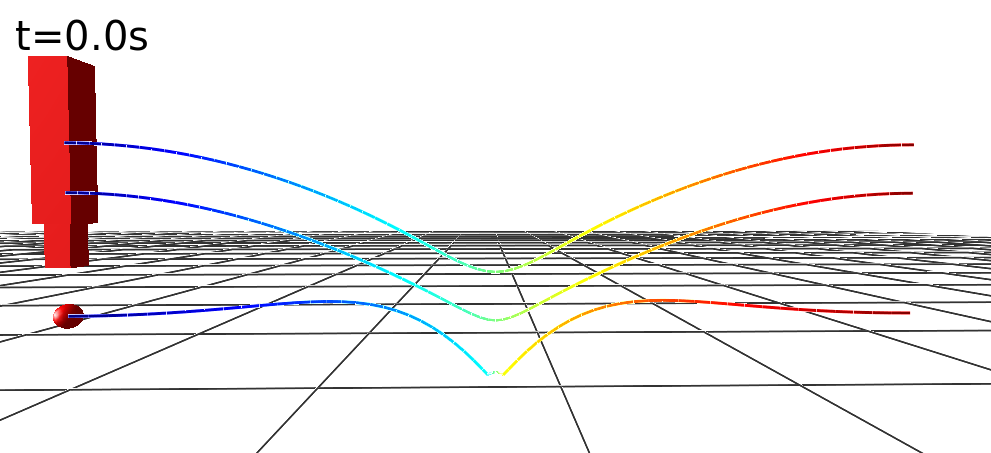}
	\includegraphics[width=0.33\textwidth]{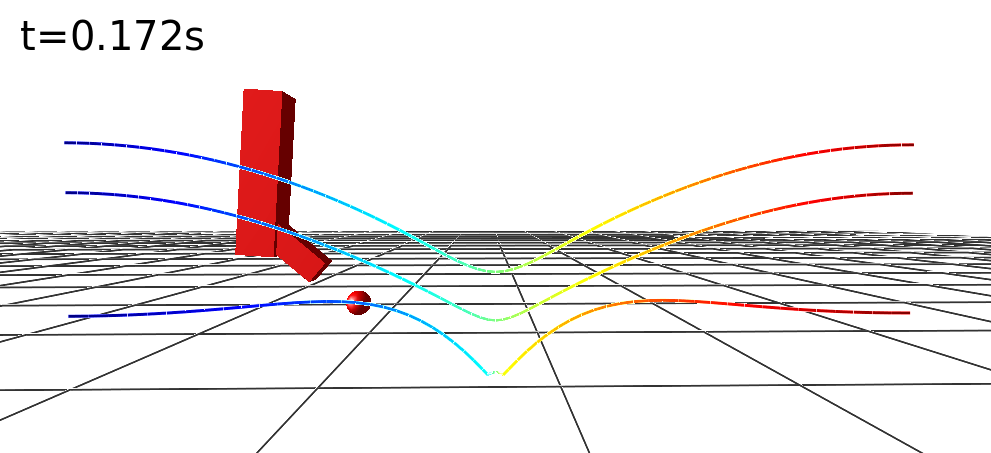}
	\includegraphics[width=0.33\textwidth]{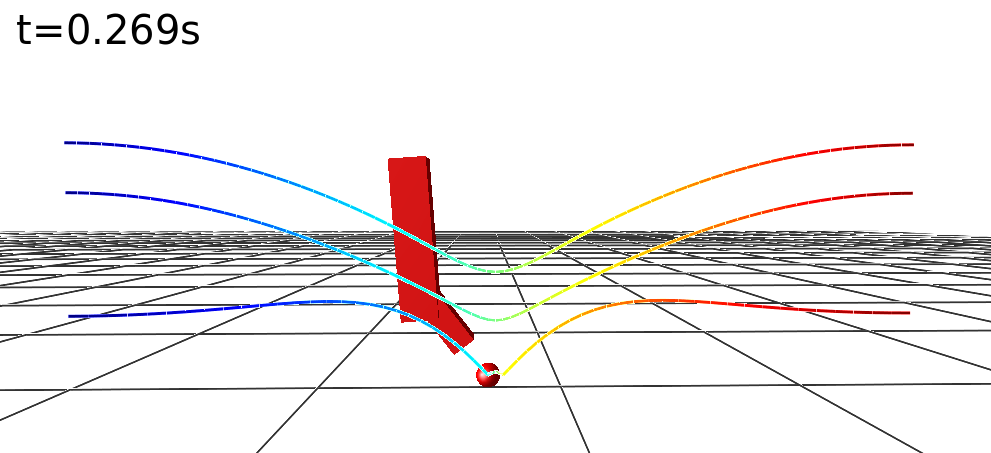}
	\includegraphics[width=0.33\textwidth]{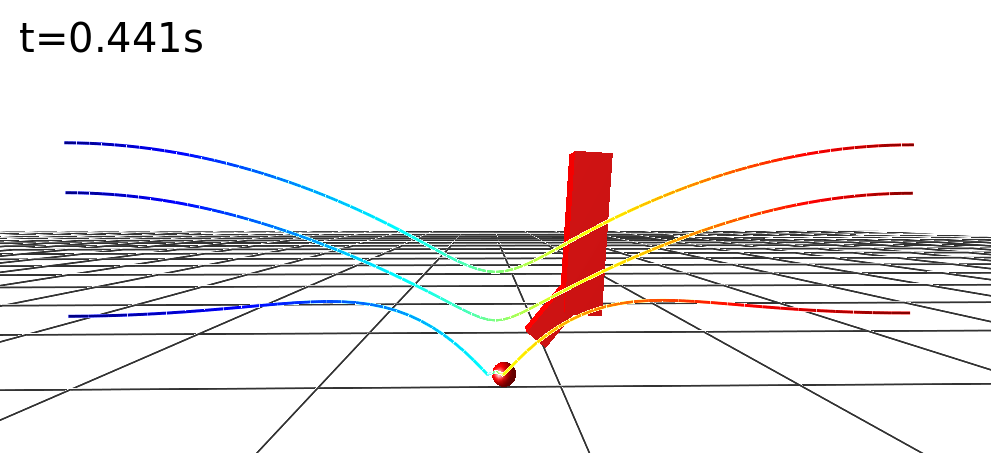}
	\includegraphics[width=0.33\textwidth]{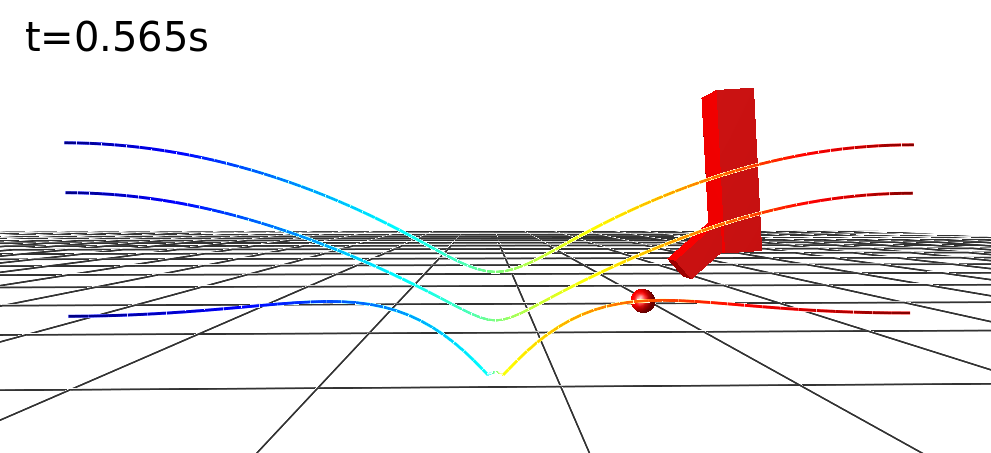}
	\includegraphics[width=0.33\textwidth]{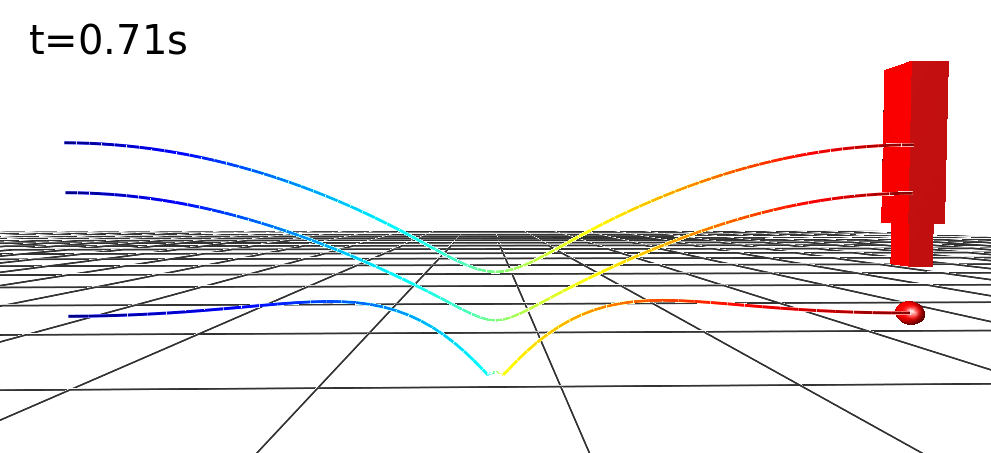}
	
	\includegraphics[width=0.33\textwidth]{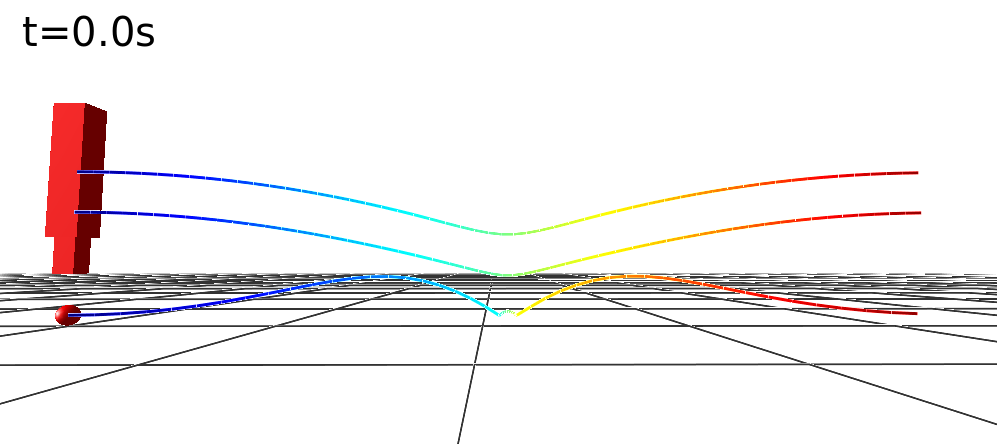}
	\includegraphics[width=0.33\textwidth]{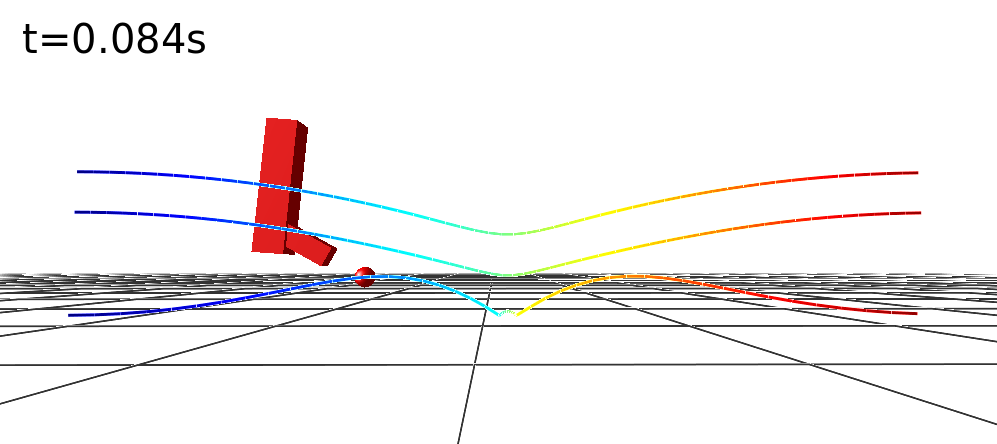}
	\includegraphics[width=0.33\textwidth]{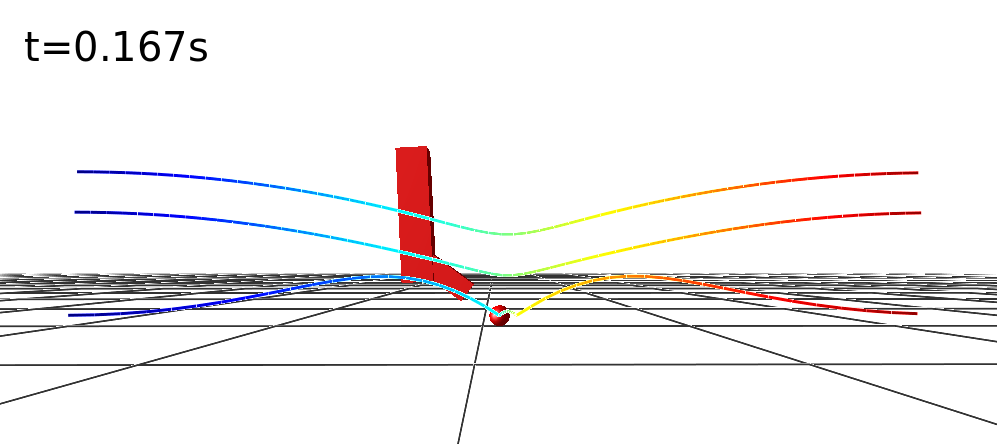}
	\includegraphics[width=0.33\textwidth]{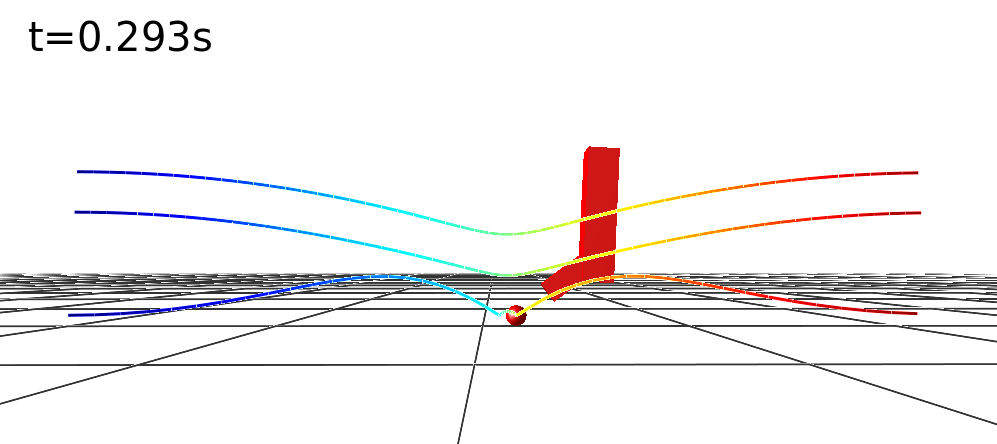}
	\includegraphics[width=0.33\textwidth]{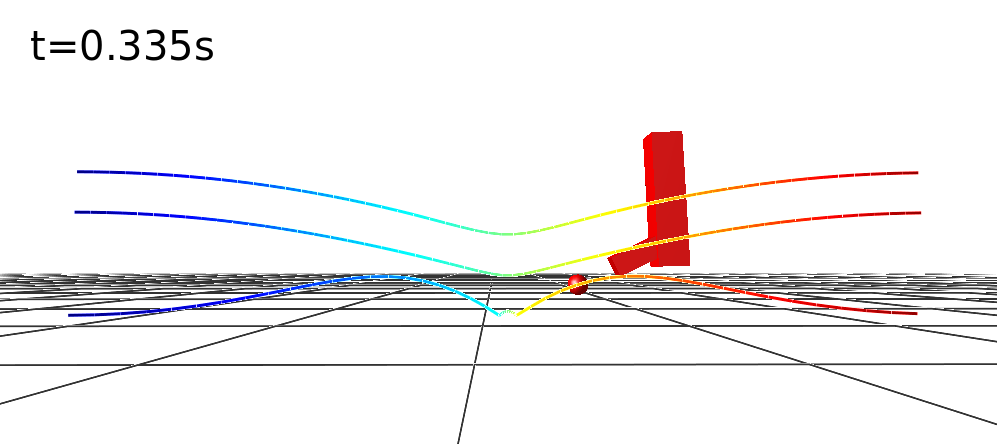}
	\includegraphics[width=0.33\textwidth]{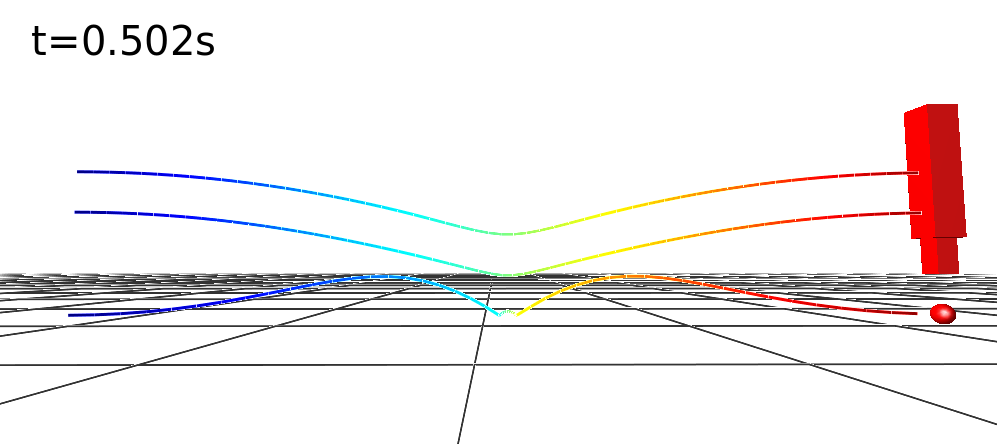}
	
	\caption{Visualization of limit cycle jumping for the upright-trunk cubic-spring (UTCS) model for $5  \mathrm{m}/\mathrm{s}$ (above) and $8 \mathrm{m}/\mathrm{s}$ (below).} \label{fig:Hip}
\end{figure*}


The search for a passive limit cycle for the UTLS model was unsucessful. No stiffness in the range from $1 \mathrm{Nm/rad}$ to $100 \mathrm{Nm/rad}$ led to a limit cycle.  However, the remaining three variations on the model described by \cite{hyon2004passive} yielded passive limit cycles: the CHCS, the UTCS, and the UTES models. In each case, the same model parameters and fixed velocity were used as in the first case with the exception of the hip location for the upright trunk models, which was displaced by a factor of $d$. The fixed points and the eigenvalues of the monodromy matrix can again be found in Tables \ref{tbl:fixpt} and \ref{tbl:eig}. All three limit cycles are unstable, indicated by the $z$ and ${\theta}$ eigenvalues greater than 1. The CHCS model simulation reached 5 steps before failure while the UTCS and UTES models reached 3 steps before failure. The energy loss over the gait cycle was found to be $-0.002$$J$, $-0.016$$J$, and $-0.016$$J$, respectively. The GRF vectors of all three simulations can be viewed in Fig. \ref{fig:GRF}. The limit cycle of the UTCS and UTES models are depicted in Fig. \ref{fig:phase}.

The UTCS model was additionally tested to determine the range of set velocities for which a stable limit cycle may be found. The lower and upper limits were found to be appr. $3.5\mathrm{m}/\mathrm{s}$ and $8.0\mathrm{m}/\mathrm{s}$, respectively. A visualization of the UTCS gaits for $5.0\mathrm{m}/\mathrm{s}$ and $8.0\mathrm{m}/\mathrm{s}$ is shown in \mbox{Fig. \ref{fig:Hip}}.

\section{Discussion}

The results of this work demonstrate that a passive dynamic limit cycle may be found for an upright-trunk template model with non-negligible leg inertia when a nonlinear progressive torsional hip spring stiffness is used. Similar results were obtained for both a cubic and an exponential spring stiffness. This finding is in agreement with our first hypothesis. To the authors' knowledge, this is the first time any passive limit cycle has been found for such a system \cite{hyon2004passive}. The results also show that the resulting ground reaction forces intersect at a central point above the center of mass similar to observations in nature \cite{Maus2010}. This confirms our second hypothesis that the passive limit cycle for the upright trunk model would demonstrate an inherent VPP-GRF-directing behavior. 

Another interesting finding was that passive limit cycles may be found for a given nonlinear hip spring parameter over a range of velocities. A similar characteristic was previously shown for the CHLS case in \cite{Hyon04}, though for a smaller range of velocities. This property may prove highly useful in the next stage when actuators and stabilization techniques are implemented, as it may allow easy adjustment of set velocity by the controller. In future work, the effects of different stiffness values on the range of feasible forward velocities and the eigenvalues of the found fixed points could be examined. 

Based on the work of \cite{hyon2004passive}, the passive limit cycle found for the centered-hip linear-spring (CHLS) model was expected. The result that the limit cycle is not stable was also expected. The passive limit cycle found for the centered-hip cubic-spring (CHNS) model was also unstable, but with slightly smaller eigenvalues for all parameters except for $\dot{\phi}$.
In agreement with the findings of \cite{hyon2004passive}, we were not able to find a limit cycle for the upright trunk template model when using linear spring stiffness (UTLS model).

A necessary condition for passive running is that energy is conserved through the gait cycle. For the current systems, the only source of energy dissipation is at the point of impact. Thus, the foot velocity should approach zero at touchdown in order to minimize the dissipation.  
Similar conclusions have been drawn in biomechanics studies of top athletes \cite{mann1984kinematic}. As indicated by the small energy loss observed in the current study, it can be concluded that this condition has not been fully met.

Given the passive limit cycle for the upright trunk model, the next step will be to explore stabilization techniques, e.g. local feedback \cite{hyon2004passive}, delayed feedback control \cite{hyon2004energy} or transversal stabilization \cite{manchester2011stable} in order to generate an efficient, stable and robust limit cycle. Toward this aim, other forms of non-linearity should be explored, such as including nonlinear or asymmetrical leg spring stiffness for improved disturbance rejection \cite{karssen2011running,xue2017motion}. Furthermore, better template matching should be considered in a next step, including aspects such as the effect of foot mass and leg retraction as well as the bipedal case. Thus, our RHM could also serve as a virtual model for control. The spring characteristics can be realized on an articulated robotic leg for example by a polar impedance controller as in \cite{Hyun2014}.

\section{Conclusion}
Using a progressively nonlinear torsional spring at the hip, a passive limit cycle was found for the first time for an upright-trunk template model with non-negligible leg mass. This is an important step toward developing simple, energy-efficient controllers for inherently unstable upright running and hopping of robots.

\section*{Acknowledment}
We gratefully acknowledge the funding support of Microsoft Germany, the Alfried Krupp von Bohlen and Halbach Foundation, and the European Union’s Horizon 2020 research and innovation programme as part of the project Darko under grant no. 101017274.

\bibliographystyle{ieeetr}
\addtolength{\textheight}{-11.5cm}
\bibliography{references}
\balance

\end{document}

%% file: intro.tex
\section{Introduction}
Upright running, or hopping, is a highly dynamic process of an inherently unstable system and it remains to be solved how humans and animals are able to accomplish such a complex task with high efficiency and minimal concentration (i.e. computational complexity). There are a number of strategies being pursued in an attempt to achieve similar feats in robotic locomotion. Reduced order and compliant control approaches, for example using template models, are one area that are explored extensively toward this aim \cite{ahmadi1997stable}.  

\begin{figure}
	\vspace{-0.5cm}
	\centering
	\def\svgwidth{0.45\textwidth}
	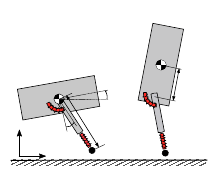 
	\caption[Model overview]{Sketch of the considered models with parameter definition. The left model corresponds to the one-legged hopping robot of \cite{Hyon04}, where the hip is coincident with the CoM ("centered-hip model"). The right model has the hip offset by the factor $d$ ("upright-trunk model").}\label{fig:Model}
\end{figure}

Blickhan's spring-loaded inverted pendulum (SLIP; \cite{Blickhan89}) and its variants remain some of the most heavily used template models for robot control. However, because the legs are considered massless in typical SLIP models, the swing phase dynamics are ignored and the swing function is limited to optimally setting the angle of attack for stable landing \cite{Blum2010,Raibert86}. Thus, important locomotion challenges associated with e.g. torso stabilization, synchronization of hip swing with vertical motion and touchdown angle requirements, as well as passive dynamics matching are ignored in these models. One noteable exception is a robot hopping template model (RHM) that is made up of a rigid trunk and leg with non-neglible moment of inertia as well as a hip and (lower) leg spring (Fig. \ref{fig:Model}, left) \cite{thompson1990passive}. This model represents the full body dynamics (including the physical interaction between the legs and the torso) during both the stance and swing phases of gait. The model can be further extended to represent an upright torso by displacing the hip from the center of mass (CoM), as shown in \cite{hyon2004passive} (Fig. \ref{fig:Model}, right). 
One of the many benefits of a template model is the ability to explore passive dynamic limit cycles. Minimizing energy requirement (or metabolic cost) is a strong motivation in robotics \cite{ahmadi1997stable} and is a fundamental principle behind animal locomotion \cite{sasaki2006muscle}. Moreover, biological locomotion is characterized by high dexterity and robustness with presumably low computational complexity, as demonstrated by the relatively low signal frequency, low control bandwidth, and high signal delay of humans vs. robots \cite{Winter1998}. Thompson and Raibert showed that, given the necessary input conditions, fully passive dynamic running could be achieved using an RHM model with the hip coincident with the CoM \cite{thompson1990passive}. Of note, the resulting limit cycle is not stable. A purely passive locomotion approach is not realistic for mechanical systems, of course, because of unavoidable mechanical losses. Rather, the idea is to tune the gait of the target robotic system using this approach, so that the passive energy return, and thus efficiency, is optimized. In fact, it has been shown that, given a passive limit cycle, active stabilization and disturbance rejection can be achieved using relatively simple controllers while exploiting the energy savings of the passive dynamics \cite{ahmadi1997stable, hyon2004passive, hyon2004energy}. However, this approach has not yet been successfully applied to an upright trunk model because no passive limit cycle could be found \cite{hyon2004passive}. 

In recent biomechanics literature \cite{Maus2010} it has been shown that an indirect trunk stabilization strategy can be observed in humans and animals. During stance, the ground reaction forces (GRFs) intersect at a point above the CoM. This has the effect that the CoM behaves like it is hanging from the intersection point, thus forming a virtual pendulum and indirectly stabilizing the upper body. The intersection point is sometimes also called the virtual pivot point (VPP).
Sharbafi et al. \cite{Sharbafi2012} approached this empirical finding by developing a VPP-based controller for trunk stabilization. However, it is unlikely that the trajectory of the ground reaction force with respect to the dynamic VPP could be adequately represented by a biological nervous system, much less be used as a control target. Thus, this empirical observation more likely points to another regulatory mechanism out of which the VPP-GRF-directing behavior emerges \cite{gruben2012force}. Identifying this regulatory mechanism could be the key to finding a passive limit cycle for the unified stance-swing template model with upright trunk. 

We hypothesized that a passive dynamic limit cycle may be found for an upright trunk model with a non-negligible leg moment of inertia by using a non-linear progressive torsional hip spring (referred to as the upright-trunk nonlinear-spring, UTNS, model). We further hypothesized that this passive limit cycle would demonstrate an inherent VPP-GRF-directing behavior of the system. Finally, as a means of comparison, we investigated whether a similar non-linear hip spring may be used to increase the stability of the same model with the hip centered at the CoM (i.e. centered-hip nonlinear-spring, CHNL, model). The main outcome and contribution of this paper is a set of RHM-based models with passive limit cycles and VPP-GRF-directing behavior. In particular, to the authors' knowledge the first passive limit cycle has been found for an upright trunk version of this model.

%% file: figures/Model_stance.pdf_tex
\begingroup%
\makeatletter%
\providecommand\color[2][]{%
	\errmessage{(Inkscape) Color is used for the text in Inkscape, but the package 'color.sty' is not loaded}%
	\renewcommand\color[2][]{}%
}%
\providecommand\transparent[1]{%
	\errmessage{(Inkscape) Transparency is used (non-zero) for the text in Inkscape, but the package 'transparent.sty' is not loaded}%
	\renewcommand\transparent[1]{}%
}%
\providecommand\rotatebox[2]{#2}%
\newcommand*\fsize{\dimexpr\f@size pt\relax}%
\newcommand*\lineheight[1]{\fontsize{\fsize}{#1\fsize}\selectfont}%
\ifx\svgwidth\undefined%
\setlength{\unitlength}{103.60831914bp}%
\ifx\svgscale\undefined%
\relax%
\else%
\setlength{\unitlength}{\unitlength * \real{\svgscale}}%
\fi%
\else%
\setlength{\unitlength}{\svgwidth}%
\fi%
\global\let\svgwidth\undefined%
\global\let\svgscale\undefined%
\makeatother%
\begin{picture}(1,0.8138595)%
\lineheight{1}%
\setlength\tabcolsep{0pt}%
\put(0,0){\includegraphics[width=\unitlength,page=1]{./figures/Model_stance.pdf}}%
\put(0.21840694,0.390806){\color[rgb]{0,0,0}\makebox(0,0)[lt]{\lineheight{1.25}\smash{\begin{tabular}[t]{l}$(x,z)$\end{tabular}}}}%
\put(0.22289947,0.26){\color[rgb]{1,0.16862745,0.14117647}\makebox(0,0)[lt]{\lineheight{1.25}\smash{\begin{tabular}[t]{l}$u_1$\end{tabular}}}}%
\put(0.4116528,0.23314111){\color[rgb]{0,0,0}\makebox(0,0)[lt]{\lineheight{1.25}\smash{\begin{tabular}[t]{l}$l$\end{tabular}}}}%

\put(0.4627851,0.357){\color[rgb]{0,0,0}\makebox(0,0)[lt]{\lineheight{1.25}\smash{\begin{tabular}[t]{l}$\theta$\end{tabular}}}}%
\put(0.30756745,0.23891757){\color[rgb]{0,0,0}\makebox(0,0)[lt]{\lineheight{1.25}\smash{\begin{tabular}[t]{l}$\phi$\end{tabular}}}}%
\put(0.02806878,0.19320545){\color[rgb]{0,0,0}\makebox(0,0)[lt]{\lineheight{1.25}\smash{\begin{tabular}[t]{l}$\vect{e}_z$\end{tabular}}}}%
\put(0.19598186,0.10802781){\color[rgb]{0,0,0}\makebox(0,0)[lt]{\lineheight{1.25}\smash{\begin{tabular}[t]{l}$\vect{e}_x$\end{tabular}}}}%

\put(0.83019779,0.40201032){\color[rgb]{0,0,0}\makebox(0,0)[lt]{\lineheight{1.25}\smash{\begin{tabular}[t]{l}$d$\end{tabular}}}}%

\put(0.33298513,0.14672726){\color[rgb]{1,0.16862745,0.14117647}\makebox(0,0)[lt]{\lineheight{1.25}\smash{\begin{tabular}[t]{l}$u_2$\end{tabular}}}}%
\put(0.76714029,0.55073148){\color[rgb]{0,0,0}\makebox(0,0)[lt]{\lineheight{1.25}\smash{\begin{tabular}[t]{l}$M$, $J_b$\end{tabular}}}}%
\put(0.75600612,0.25421783){\color[rgb]{0,0,0}\makebox(0,0)[lt]{\lineheight{1.25}\smash{\begin{tabular}[t]{l}$J_l$\end{tabular}}}}%
\end{picture}%
\endgroup%

%% file: methods.tex
\section{Methods}	
\subsection{Modeling} 
\textbf{Robot Dynamics.} A legged robot can be modeled as a floating-base system \cite{Nenchev2018}. The configuration of the robot is denoted $\vect{q} = \left[\vect{q}_b, \vect{q}_j\right]^{\mathsf{T}} \in \mathbb{R}^n$, where $\vect{q}_b \in \mathbb{R}^{n_b}$ represents the floating base posture, $\vect{q}_j \in \mathbb{R}^{n_j}$ denotes the joint configuration and $n = n_b + n_j$ is the total number of DoFs.

The dynamics of the robot can be stated as
\begin{equation}
\mat{M}(\vect{q}) \vect{\ddot{q}} + \vect{h}(\vect{q}, \vect{\dot{q}}) = \vect{S}^{\mathsf{T}}\vect{u} \label{eq:free-float}
\end{equation}
where $\mat{M}(\vect{q}) \in \mathbb{R}^{n \times n}$ represents the mass matrix, $\vect{h}(\vect{q}, \vect{\dot{q}}) \in \mathbb{R}^{n \times 1}$ the nonlinear bias vector including Coriolis, centrifugal and gravitational forces, $\vect{u} \in \mathbb{R}^{m \times 1}$ the vector of generalized joint forces and $\mat{S}$ is a matrix that selects the actuated DoFs. 

In case the foot of the robot is in contact with the ground, one has to consider the constrained dynamics of a floating-base robot, which can be stated as  \\
	\begin{align}
		\mat{M}(\vect{q}) \vect{\ddot{q}} + \vect{h}(\vect{q}, \vect{\dot{q}}) &= \vect{S}^{\mathsf{T}}\vect{u} + \vect{J}^{\mathsf{T}} \vect{\lambda} \\
		 \vect{c}(\vect{q}) &= \vect{0}, \label{eq:constr}
\end{align}
where $\vect{c}(\vect{q})$ is the position of the foot. The transpose of the contact Jacobian $\vect{J}=\frac{\partial \vect{c}(\vect{q})}{\partial{\vect{q}}} \in \mathbb{R}^{n_c \times n}$ maps the contact forces $\vect{\lambda} \in \mathbb{R}^{n_c}$ to the robot dynamics. The contact condition Eq. \eqref{eq:constr} can be expressed equivalently in differential form
\begin{align}
 \vect{\dot{c}}(\vect{q}) &= \vect{J} \vect{\dot{q}} = \vect{0}\\
 \vect{\ddot{c}}(\vect{q}) &= \mat{J} \vect{\ddot{q}} + \vect{\dot{J}} \vect{\dot{q}} = \vect{0}
\end{align}
which leads to the following system of equations:
\begin{equation}
	\left[\begin{array}{cc}
	\boldsymbol{M}(\boldsymbol{q}) & \boldsymbol{J}(\boldsymbol{q})^{\mathsf{T}} \\
	\boldsymbol{J}(\boldsymbol{q}) & \boldsymbol{0}
	\end{array}\right]\left[\begin{array}{c}
	\ddot{\boldsymbol{q}} \\
	-\boldsymbol{\lambda}
	\end{array}\right]=\left[\begin{array}{c}
	-\boldsymbol{h}(\boldsymbol{q}, \dot{\boldsymbol{q}})+\vect{S}^{\mathsf{T}}\boldsymbol{u} \\
	\boldsymbol{\gamma}(\boldsymbol{q}, \dot{\boldsymbol{q}})
	\end{array}\right] \label{eq:constrained_system}
	\end{equation}
Here, $\boldsymbol{\gamma}(\boldsymbol{q}, \dot{\boldsymbol{q}}) = \mat{J} \vect{\ddot{q}}$ represents the contact Hessian \cite{Felis2016}. \\
\\
\textbf{Collision.} Collision occurs when the robot's foot impacts the ground. We make the assumptions that the collision is inelastic, the duration of the collision is infinitesimally small and friction is sufficiently high such that no slipping occurs. At the time of the collision, conservation of momentum yields
\begin{equation}
\int_{t_{c}^{-}}^{t_{c}^{+}} \boldsymbol{M}(\boldsymbol{q}) \ddot{\boldsymbol{q}}+\boldsymbol{h}(\boldsymbol{q}, \dot{\boldsymbol{q}}) \mathrm{d} t=\int_{t_{c}^{-}}^{t_{c}^{+}} \boldsymbol{u}+\boldsymbol{J}(\boldsymbol{q})^{T} \boldsymbol{\lambda} \mathrm{d} t. \label{eq:coll_full}
\end{equation}
Since we assumed that the impact duration is infinitesimally small, we define $t_c^{+} = \underset{t \rightarrow t_c}{\lim \inf }\text{\;} t$, and $t_c^{-} = \underset{t \rightarrow t_c}{\lim \sup }\text{\;} t$ \cite{Sobotka2007}. No discontinuities in the joint position are allowed. This means $\vect{q}(t_c^{+}) = \vect{q}(t_c^{-})$ holds, and thus we can evaluate the integrals in Eq. \eqref{eq:coll_full}, yielding
\begin{equation}
\boldsymbol{M}(\boldsymbol{q})\left(\dot{\boldsymbol{q}}\left(t_{c}^{+}\right)-\dot{\boldsymbol{q}}\left(t_{c}^{-}\right)\right)=\boldsymbol{J}(\boldsymbol{q})^{\mathsf{T}} \vect{\Lambda}, \label{eq:coll_eval}
\end{equation}
where \begin{equation}
	\vect{\Lambda}=\int_{t_{c}^{-}}^{t_{c}^{+}} \boldsymbol{\lambda} \mathrm{d} t
\end{equation}
 is the contact impulse. Equivalently, Eq. \eqref{eq:coll_eval} can be expressed as

\begin{equation} \label{eq:impact}
\left[\begin{array}{cc}
\boldsymbol{M}(\boldsymbol{q}) & \boldsymbol{J}(\boldsymbol{q})^{\mathsf{T}} \\
\boldsymbol{J}(\boldsymbol{q}) & \boldsymbol{0}
\end{array}\right]\left[\begin{array}{c}
\vect{\dot{q}}(t_c^{+}) \\
-\boldsymbol{\Lambda}
\end{array}\right]=\left[\begin{array}{c}
\boldsymbol{M}(\boldsymbol{q})\vect{\dot{q}}(t_c^{-}) \\
-e \boldsymbol{J}(\boldsymbol{q}) \vect{\dot{q}}(t_c^{-})
\end{array}\right]
\end{equation}
Due to the inelasticity assumption, no rebound is allowed, and hence $e = 0$.\\
\\
\textbf{Detachment.} When the foot detaches from the ground, it has no effect on the velocities. Therefore, for the velocity before and after the impact 
\begin{equation}
	\vect{\dot{q}}(t_c^{-}) = \vect{\dot{q}}(t_c^{+}) \label{eq:detach}
\end{equation} holds.\\
\\
\textbf{Hybrid system model.} During locomotion, a legged robot exhibits a sequence of different contact situations, where the robot gains or loses contact with the environment. Depending on the nature of contact situation, the dynamics are either governed by Eq. \eqref{eq:free-float} or Eq. \eqref{eq:constrained_system}. The modeling approach described above allows for instantaneous discontinuities in the velocities when an impact occurs. A hybrid system model that guards the transitions between the phases is therefore a well-suited approach to describe the complete gait cycle. \\
A hybrid system model is comprised of the continuous dynamics for each phase \cite{hyon2004passive}
\begin{equation}
				\vect{\dot{x}}_i = \vect{f}_i(\vect{x}_i), 
\end{equation}
which corresponds to either Eq. \eqref{eq:free-float} or Eq. \eqref{eq:constr} with the state vector $\vect{x} = \left[\vect{q},\vect{\dot{q}}\right]^{\mathsf{T}}$, a jump map
\begin{equation}
	\vect{x}_i^{+} = \vect{g}_i(\vect{x}_i^{-}),
\end{equation}
which is evaluated according to either Eq. \eqref{eq:impact} or Eq. \eqref{eq:detach} and discrete events or cross sections
\begin{equation}
	\Sigma_i  \coloneqq \left\lbrace\vect{x}_i \mid s_i(\vect{x}) = 0 \right\rbrace
\end{equation}
that are governed by transition surfaces $s_i(\vect{x})$. Starting at a certain discrete event, the hybrid trajectory is obtained by consecutive time-integration and evaluation of the jump maps for each phase. In this context, a periodic orbit means that at the last event of the hybrid trajectory the jump map again maps to the initial value. Fig. \ref{fig:poincare} illustrates this idea.
						
\subsection{Application to one-legged robots}

In this paper, the focus lies on two specific cases of one-legged hopping robots with 1) the hip centered at the CoM ("centered-hip model") and 2) an offset hip and an upright trunk ("upright trunk model"; Fig. \ref{fig:Model}). The robots each have five DoFs \mbox{$\vect{q} = \left[x, z, \theta, \phi, l\right]^{\mathsf{T}}$}. Each robot consists of a rigid trunk and upper leg segment, a hip and a leg spring, and a point foot. As in \cite{hyon2004passive, ahmadi1997stable}, the point foot is assumed to be massless. 
We are interested in passive gaits, hence there are no actuator inputs; the only forces and torques acting on the joints result from springs, as described by
\begin{table}
	\vspace{0.2cm}
	\caption{Model and control parameter values. The mechanical parameters were chosen similar to \cite{hyon2004energy}. The hip spring stiffness values were found in the gait search algorithm.}\label{tbl:values}
	\vspace{0.25cm}
	\begin{tabularx}{0.47\textwidth}{c|c|c|c|c|c|c}
		
		Par. & Units & CHLS & CHCS & UTLS & UTCS & UTES \\
		\hline  
		& & & & & & \\
		$M$ & $\mathrm{kg}$ & $12$  & $12$  & $12$  & $12$ & $12$ \\ 
		$l_0$ & $\mathrm{m}$ & $0.5$  & $0.5$  & $0.5$ & $0.5$ & $0.5$ \\ 
		$J_b$ & $\mathrm{kgm^2}$ & $0.5$  & $0.5$  & $0.5$ & $0.5$ & $0.5$ \\ 
		$J_l$ & $\mathrm{kgm^2}$ & $0.11$  & $0.11$  & $0.11$ & $0.11$ & $0.11$ \\ 
		$d$ & $\mathrm{m}$ & $0.0$  & $0.0$  & $0.2$ & $0.2$ & $0.2$ \\ 
		$K_l$ & $\mathrm{N/m}$ & $3000$  & $3000$  & $3000$ & $3000$ & $3000$ \\ 
		$K_{h}$ & $\mathrm{Nm/rad}$ & $10.0$  & $19.98$ & $-$ & $10$ & $2$ \\  
		
	\end{tabularx} 
	\vspace{-0.25cm}
	
\end{table}
\begin{equation}
\vect{u} = -\frac{\partial\vect{U}(\vect{q},\vect{q_0})}{\partial \vect{q}},
\end{equation}
where $\vect{U}$ is the spring potential. All robots are equipped with a linear leg spring
\begin{equation}
u_2 = K_l (l_0 - l).
\end{equation}
Each robot is tested using either two classes of hip spring. The first is a linear spring, where
\begin{equation}
u_1 = K_{h} (\phi_0 - \phi).
\end{equation}
This leads to the centered-hip linear-spring (CHLS) and upright-trunk linear-spring (UTLS) models.
The second type of spring is a nonlinear progressive spring. We considered both a cubic spring
\begin{equation}
u_1 = K_{h} (\phi_0 - \phi)^3,
\end{equation}
and an exponential spring
\begin{equation}
u_1 = K_{h} (e^{(\phi_0 - \phi)} - e^{-(\phi_0 - \phi)})  .
\end{equation}

This leads to the centered-hip cubic-spring (CHCS) as well as the upright-trunk cubic-spring and exponential-spring (UTCS and UTES) models. The full set of robot parameters are summarized in Table \ref{tbl:values}. To compare our upright hopping robot to similar hopping robots in the literature, we have chosen the same parameter set as Hyon et al. \cite{hyon2004energy} for masses and inertias. It is important to mention that $J_l$ represents an equivalent leg inertia. For simplicity, it is assumed that the leg mass is lumped with the upper body mass forming the total mass $M$. Due to the foot being massless, there is no impulse along the leg axis. However, since there is an associated equivalent leg inertia, an impulse in the perpendicular direction is still possible and can lead to an energy loss upon impact. Since there is only one foot that can collide with the ground, only two gait phases exist: The stance phase, where the point foot is in contact with the ground, and the flight phase, where there is no contact. Due to the massless foot, during the flight phase $l = l_0$ and $\dot{l} = 0$ holds.

\textbf{Discrete events.} The touchdown event occurs when the point foot touches ground
\begin{equation}
s_{\mathrm{td}}(\vect{x}) = z_{foot}.
\end{equation}
The jump map $\vect{g}_{\mathrm{lo}}$ corresponds to Eq. \eqref{eq:impact}. The liftoff event takes place when the vertical component of the GRF becomes zero
\begin{equation}
s_{\mathrm{lo}}(\vect{x}) = f_z.
\end{equation}
There are no discontinuities for the liftoff event. For subsequent use in the gait search algorithm, we define an additional transition surface at the apex
\begin{equation}
s_{\mathrm{ap}}(\vect{x}) = \dot{z},
\end{equation}
which is not associated with a phase change.
\subsection{Gait search algorithm}
In order to find periodic gaits, we employ Newton's method for Poincar\'{e} maps according to \cite{Waugh2013, Hyon04}. A Poincar\'{e} map is a mapping from a cross section  transversal to the system flow onto itself $P: \Sigma \rightarrow \Sigma$. A point $\vect{x}^* \in \Sigma$ is an element of a periodic orbit if it is a fixed point of the Poincar\'{e} map, i.e. $\vect{P}(\vect{x}^*) = \vect{x}^*$. Thus, for a fixed point 
\begin{equation}
	\vect{F}(\vect{x}) = \vect{x} - \vect{P}(\vect{x}) = \vect{0} \label{eq:poincare}
\end{equation}
must hold. This must also be fullfilled for a small variation with respect to the reference point \cite{ChaosBook}, i.e. \mbox{$\vect{F}(\vect{x} + \vect{\Delta x}) = \vect{F}(\vect{x}) = \vect{0}$}. Using a Taylor series expansion \cite{hyon2004passive}, we can write 
\begin{equation}
	\vect{F}(\vect{x} + \vect{\Delta x}) = \vect{F}(\vect{x}) + \vect{DF}(\vect{x}) \vect{\Delta x}+ \vect{O}(\vect{x}), \label{eq:Fx}
\end{equation}
where $\vect{O}(\vect{x})$ are the higher order terms (which can be neglected) and 
\begin{equation}
	\vect{DF}(\vect{x}) = \frac{\partial \vect{F}(\vect{x})}{\partial \vect{x}} = \vect{I}-\vect{DP}(\vect{x}). 
\end{equation}
Here, $\vect{DP}(\vect{x}) = \frac{\partial \vect{P}(\vect{x})}{\partial \vect{x}}$ is the so called monodromy matrix. The eigenvalues of the monodromy matrix indicate whether the limit cycle is stable or not. If all eigenvalues lie within the unit cycle, the limit cycle is stable, otherwise unstable. To find a periodic gait, Eq. \eqref{eq:Fx} can be reformulated as an iterative formula
\begin{equation}
		\left[\mat{I} - \vect{DP}(\vect{\vect{x}_k}) \right] \vect{\Delta x} =  \vect{P}(\vect{x}_k) - \vect{x}_k. 
	\end{equation}
The update step is $\vect{x}_{k+1} = \vect{x}_k + \rho \vect{\Delta x}$. Newton's method is stable only within a linear neighbourhood of the solution and thus requires a precise initial guess. To enlarge the basin of attraction, a relaxation factor $0 \leq \rho \leq 1$ can be used. We define the apex event as the cross section. Since $\dot{z} = 0$, $l = l_0$ and $\dot{l} = 0$ holds at the apex and the $x$ position is not a cyclic variable, the dimensionality of the search space is reduced by four, yielding the reduced state vector $\vect{x}_{red} = [z, \theta, \phi, \dot{x}, \dot{\theta}, \dot{\phi}]^{\mathsf{T}}$. Additionally, if we are interested in gaits for a specific forward velocity, this state can be fixed as well. 
\begin{figure}
\centering
\def\svgwidth{0.5\textwidth}
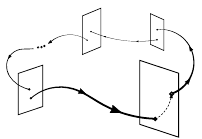 
\caption[poincare]{Illustration of a hybrid periodic orbit $\vect{\Gamma}$. The solution is comprised
	of the individual contributions of each phase $\vect{x}_i$. At each cross-section $\Sigma_i$, the jump maps $\vect{g}_i(\vect{x}_i)$ are evaluated.}\label{fig:sketch}
\end{figure}

\subsection{Software}
To model the robot dynamics with the different contact situations, we employ the Rigid Body Dynamics library (RBDL) developed by Felis \cite{Felis2016RBDL}. For  time-integration we use the Arkode solver from the Sundials framework \cite{Sundials}, with an absolute tolerance of $\epsilon_\mathrm{abs} = 10^{-10}$ and a relative tolerance $\epsilon_\mathrm{rel} = 10^{-8}$. The monodromy matrix is computed by finite differences with a precision of $\epsilon = 10^{-6}$. 

%% file: figures/poincare.pdf_tex
\begingroup%
  \makeatletter%
  \providecommand\color[2][]{%
    \errmessage{(Inkscape) Color is used for the text in Inkscape, but the package 'color.sty' is not loaded}%
    \renewcommand\color[2][]{}%
  }%
  \providecommand\transparent[1]{%
    \errmessage{(Inkscape) Transparency is used (non-zero) for the text in Inkscape, but the package 'transparent.sty' is not loaded}%
    \renewcommand\transparent[1]{}%
  }%
  \providecommand\rotatebox[2]{#2}%
  \newcommand*\fsize{\dimexpr\f@size pt\relax}%
  \newcommand*\lineheight[1]{\fontsize{\fsize}{#1\fsize}\selectfont}%
  \ifx\svgwidth\undefined%
    \setlength{\unitlength}{96.65110791bp}%
    \ifx\svgscale\undefined%
      \relax%
    \else%
      \setlength{\unitlength}{\unitlength * \real{\svgscale}}%
    \fi%
  \else%
    \setlength{\unitlength}{\svgwidth}%
  \fi%
  \global\let\svgwidth\undefined%
  \global\let\svgscale\undefined%
  \makeatother%
  \begin{picture}(1,0.6870663)%
    \lineheight{1}%
    \setlength\tabcolsep{0pt}%
    \put(0,0){\includegraphics[width=\unitlength,page=1]{./figures/poincare.pdf}}%
    \put(0.71111419,0.32785756){\color[rgb]{0,0,0}\makebox(0,0)[lt]{\lineheight{1.25}\smash{\begin{tabular}[t]{l}$\Sigma_1$\end{tabular}}}}%
    \put(0.70847748,0.17258761){\color[rgb]{0,0,0}\makebox(0,0)[lt]{\lineheight{1.25}\smash{\begin{tabular}[t]{l}$\vect{g}_1(\vect{x}_1^-)$\end{tabular}}}}%
    \put(0.75899921,0.56868357){\color[rgb]{0,0,0}\makebox(0,0)[lt]{\lineheight{1.25}\smash{\begin{tabular}[t]{l}$\Sigma_2$\end{tabular}}}}%
    \put(0.42180278,0.57946324){\color[rgb]{0,0,0}\makebox(0,0)[lt]{\lineheight{1.25}\smash{\begin{tabular}[t]{l}$\Sigma_3$\end{tabular}}}}%
    \put(0.10668345,0.2879212){\color[rgb]{0,0,0}\makebox(0,0)[lt]{\lineheight{1.25}\smash{\begin{tabular}[t]{l}$\Sigma_N$\end{tabular}}}}%
    \put(0.81470715,0.25081542){\color[rgb]{0,0,0}\makebox(0,0)[lt]{\lineheight{1.25}\smash{\begin{tabular}[t]{l}$\vect{x}_1^+$\end{tabular}}}}%
    \put(0.77025572,0.06337065){\color[rgb]{0,0,0}\makebox(0,0)[lt]{\lineheight{1.25}\smash{\begin{tabular}[t]{l}$\vect{x}_1^-$\end{tabular}}}}%
    \put(0.40494393,0.15640429){\color[rgb]{0,0,0}\makebox(0,0)[lt]{\lineheight{1.25}\smash{\begin{tabular}[t]{l}$\vect{\Gamma}$\end{tabular}}}}%
    \put(0.89286392,0.36645565){\color[rgb]{0,0,0}\makebox(0,0)[lt]{\lineheight{1.25}\smash{\begin{tabular}[t]{l}$\vect{x}_1$\end{tabular}}}}%
  \end{picture}%
\endgroup%

%% file: ICRA2021_Ossadnik.bbl
\begin{thebibliography}{10}

\bibitem{ahmadi1997stable}
M.~Ahmadi and M.~Buehler, ``Stable control of a simulated one-legged running
  robot with hip and leg compliance,'' {\em IEEE transactions on robotics and
  automation}, vol.~13, no.~1, pp.~96--104, 1997.

\bibitem{Blickhan89}
R.~Blickhan, ``The spring-mass model for running and hopping,'' {\em Journal of
  biomechanics}, vol.~22, no.~11-12, pp.~1217--1227, 1989.

\bibitem{Blum2010}
Y.~Blum, S.~W. Lipfert, J.~Rummel, and A.~Seyfarth, ``Swing leg control in
  human running,'' {\em Bioinspiration {\&} Biomimetics}, vol.~5, p.~026006,
  may 2010.

\bibitem{Raibert86}
M.~H. Raibert, {\em Legged robots that balance}.
\newblock MIT press, 1986.

\bibitem{thompson1990passive}
C.~M. Thompson and M.~H. Raibert, ``Passive dynamic running,'' in {\em
  Experimental Robotics I}, pp.~74--83, Springer, 1990.

\bibitem{hyon2004passive}
S.-H. Hyon, X.~Jiang, T.~Emura, and T.~Ueta, ``Passive running of planar
  1/2/4-legged robots,'' in {\em 2004 IEEE/RSJ International Conference on
  Intelligent Robots and Systems (IROS)(IEEE Cat. No. 04CH37566)}, vol.~4,
  pp.~3532--3539, IEEE, 2004.

\bibitem{Hyon04}
S.-H. Hyon, T.~Emura, and T.~Ueta, ``Detection and stabilization of hybrid
  periodic orbits of passive running robots,'' in {\em In Mechatronics and
  Robotics}, p.~1314, 2004.

\bibitem{sasaki2006muscle}
K.~Sasaki and R.~R. Neptune, ``Muscle mechanical work and elastic energy
  utilization during walking and running near the preferred gait transition
  speed,'' {\em Gait \& posture}, vol.~23, no.~3, pp.~383--390, 2006.

\bibitem{Winter1998}
D.~A. Winter, A.~E. Patla, F.~Prince, M.~Ishac, and K.~Gielo-Perczak,
  ``Stiffness control of balance in quiet standing,'' {\em Journal of
  Neurophysiology}, vol.~80, no.~3, pp.~1211--1221, 1998.
\newblock PMID: 9744933.

\bibitem{hyon2004energy}
S.-H. Hyon and T.~Emura, ``Energy-preserving control of a passive one-legged
  running robot,'' {\em Advanced Robotics}, vol.~18, no.~4, pp.~357--381, 2004.

\bibitem{Maus2010}
H.-M. Maus, S.~Lipfert, M.~Gross, J.~Rummel, and A.~Seyfarth, ``Upright human
  gait did not provide a major mechanical challenge for our ancestors,'' {\em
  Nature communications}, vol.~1, p.~70, 2010.

\bibitem{Sharbafi2012}
M.~A. {Sharbafi}, C.~{Maufroy}, H.~M. {Maus}, A.~{Seyfarth}, M.~N.
  {Ahmadabadi}, and M.~J. {Yazdanpanah}, ``Controllers for robust hopping with
  upright trunk based on the virtual pendulum concept,'' in {\em 2012 IEEE/RSJ
  International Conference on Intelligent Robots and Systems}, pp.~2222--2227,
  Oct 2012.

\bibitem{gruben2012force}
K.~G. Gruben and W.~L. Boehm, ``Force direction pattern stabilizes sagittal
  plane mechanics of human walking,'' {\em Human movement science}, vol.~31,
  no.~3, pp.~649--659, 2012.

\bibitem{Nenchev2018}
D.~N. Nenchev, A.~Konno, and T.~Tsujita, {\em Humanoid robots: Modeling and
  control}.
\newblock Butterworth-Heinemann, 2018.

\bibitem{Felis2016}
M.~L. {Felis} and K.~{Mombaur}, ``Synthesis of full-body 3-{D} human gait using
  optimal control methods,'' in {\em 2016 IEEE International Conference on
  Robotics and Automation (ICRA)}, pp.~1560--1566, 2016.

\bibitem{Sobotka2007}
M.~Sobotka, {\em Hybrid dynamical system methods for legged robot locomotion
  with variable ground contact}.
\newblock PhD thesis, Technische Universit{\"a}t M{\"u}nchen, 2007.

\bibitem{Waugh2013}
I.~Waugh, S.~Illingworth, and M.~Juniper, ``Matrix-free continuation of limit
  cycles for bifurcation analysis of large thermoacoustic systems,'' {\em
  Journal of Computational Physics}, vol.~240, pp.~225 -- 247, 2013.

\bibitem{ChaosBook}
P.~Cvitanovi{\'c}, R.~Artuso, R.~Mainieri, G.~Tanner, and G.~Vattay, {\em
  Chaos: Classical and Quantum}.
\newblock Copenhagen: Niels Bohr Inst., 2016.

\bibitem{Felis2016RBDL}
M.~L. Felis, ``{RBDL}: {A}n efficient rigid-body dynamics library using
  recursive algorithms,'' {\em Autonomous Robots}, pp.~1--17, 2016.

\bibitem{Sundials}
A.~C. Hindmarsh, P.~N. Brown, K.~E. Grant, S.~L. Lee, R.~Serban, D.~E.
  Shumaker, and C.~S. Woodward, ``{SUNDIALS}: Suite of nonlinear and
  differential/algebraic equation solvers,'' {\em ACM Transactions on
  Mathematical Software (TOMS)}, vol.~31, no.~3, pp.~363--396, 2005.

\bibitem{mann1984kinematic}
R.~Mann, J.~Kotmel, J.~Herman, B.~Johnson, and C.~Schultz, ``Kinematic trends
  in elite sprinters,'' in {\em ISBS-Conference Proceedings Archive}, 1984.

\bibitem{manchester2011stable}
I.~R. Manchester, U.~Mettin, F.~Iida, and R.~Tedrake, ``Stable dynamic walking
  over uneven terrain,'' {\em The International Journal of Robotics Research},
  vol.~30, no.~3, pp.~265--279, 2011.

\bibitem{karssen2011running}
J.~D. Karssen and M.~Wisse, ``Running with improved disturbance rejection by
  using non-linear leg springs,'' {\em The International Journal of Robotics
  Research}, vol.~30, no.~13, pp.~1585--1595, 2011.

\bibitem{xue2017motion}
T.~Xue, J.~Zhao, and J.~Wang, ``Motion control for variable stiffness slip
  model of legged robot single leg,'' in {\em 2017 Chinese Automation Congress
  (CAC)}, pp.~4711--4716, IEEE, 2017.

\bibitem{Hyun2014}
D.~J. Hyun, S.~Seok, J.~Lee, and S.~Kim, ``High speed trot-running:
  Implementation of a hierarchical controller using proprioceptive impedance
  control on the mit cheetah,'' {\em The International Journal of Robotics
  Research}, vol.~33, no.~11, pp.~1417--1445, 2014.

\end{thebibliography}
